\documentclass[letterpaper]{article} 
\usepackage{aaai25}  
\usepackage{times}  
\usepackage{helvet}  
\usepackage{courier}  
\usepackage[hyphens]{url}  
\usepackage{graphicx} 
\usepackage{natbib}  
\usepackage{caption} 

\usepackage[latin9]{inputenc}
\usepackage{array}
\usepackage{float}
\usepackage{textcomp}
\usepackage{multirow}
\usepackage{amsmath}
\usepackage{graphicx}
\providecommand{\tabularnewline}{\\}

\usepackage{algorithm}
\usepackage{algorithmic}

\usepackage{newfloat}
\usepackage{listings}
\DeclareCaptionStyle{ruled}{labelfont=normalfont,labelsep=colon,strut=off} 
\floatstyle{ruled}
\newfloat{listing}{tb}{lst}{}
\floatname{listing}{Listing}
\title{GBRIP: Granular Ball Representation for Imbalanced Partial Label Learning}
\author{
    Jintao Huang\textsuperscript{\rm 1},
    Yiu-ming Cheung\textsuperscript{\rm 1,}\thanks{Corresponding author.},
    Chi-man Vong\textsuperscript{\rm 2},
    Wenbin Qian\textsuperscript{\rm 3}
}
\affiliations{
    \textsuperscript{\rm 1}Department of Computer Science, Hong Kong Baptist University, Hong Kong SAR, China\\
    \textsuperscript{\rm 2}Department of Computer and Information Science, University of Macau, Macau SAR, China\\
    \textsuperscript{\rm 3}School of Software, Jiangxi Agricultural University, Nanchang, China\\
    \{csjthuang,ymc\}@comp.hkbu.edu.hk; cmvong@um.edu.mo; wenbinqian1027@126.com
}

\usepackage{bibentry}

\begin{document}

\maketitle

\begin{abstract}
Partial label learning (PLL) is a complicated weakly supervised multi-classification
task compounded by class imbalance. Currently, existing methods only rely on inter-class
pseudo-labeling from inter-class features, often overlooking the significant impact of the intra-class imbalanced features combined with the inter-class. To address these
limitations, we introduce Granular Ball Representation for Imbalanced PLL (GBRIP), a novel framework for imbalanced PLL.
GBRIP utilizes coarse-grained granular ball representation and multi-center
loss to construct a granular ball-based feature space through unsupervised
learning, effectively capturing the feature distribution within each
class. GBRIP mitigates the impact of confusing features by systematically
refining label disambiguation and estimating imbalance distributions.
The novel multi-center loss function enhances learning by emphasizing
the relationships between samples and their respective centers within
the granular balls. Extensive experiments on standard benchmarks demonstrate
that GBRIP outperforms existing state-of-the-art methods, offering
a robust solution to the challenges of imbalanced PLL. The code will be available in supplemental materials.
\end{abstract}

%

\section{Introduction}

Partial Label Learning (PLL) (\citet{cour2011learning,tian2023partial,10416802,xu2025label})
has emerged as a crucial area in weakly supervised learning, addressing
challenges in image annotation, natural language processing, and web
mining. PLL scenarios involve instances associated with multiple candidate
labels, with only one correct label mirroring real-world labeling
complexities (\citet{hullermeier2006learning,cour2011learning,chen2017learning,tian2023partial}).
Recent research has progressed from fundamental techniques to more
advanced methods, which can be categorized into two approaches: Average
Disambiguation Method (ADM) and Identification Disambiguation Method
(IDM) (\citet{tian2023partial}). ADM, utilized in models such as
Partial Label k-Nearest Neighbors (PL-kNN) (\citet{hullermeier2006learning})
and Partial Label Support Vector Machine (PL-SVM) (\citet{nguyen2008classification}),
averages scores across all candidate labels, potentially influenced
by incorrect labels. On the other hand, IDM treats the true label
as a latent variable and employs iterative methods like Maximum Likelihood
Estimation (\citet{liu2012conditional,lv2020progressive}). Nonetheless,
it runs the risk of misidentifying false positives as true labels.
Recent advancements in Graph-based manifold learning (\citet{wang2019adaptive,lyu2022deep,zhang2020learning,10539299})
have enhanced PLL by capitalizing on data relationships and intrinsic
geometries, improving accuracy and robustness.

Numerous PLL methods are designed to assume balanced class distributions,
which do not always align with real-world data that often exhibit an imbalanced distribution \citet{Lu_2023_ICCV,10491302}. This imbalance can lead to suboptimal performance
in less classes and cause predictions to be biased toward
dominant categories. While addressing class imbalance in multi-class
classification is well-researched, it presents unique challenges in
PLL inexact labeling (\citet{zhang2023deep}).
Due to label ambiguity, traditional methods such as under- or over-sampling
could not be effective in PLL. Imbalanced PLL (IPLL) (\citet{wang2018towards})
and Long-tailed PLL (LT-PLL) are more viable (\citet{wang2022solar}),
but existing PLL methods struggle in  labels disambiguating in imbalanced
settings. Innovative approaches are necessary to tackle class imbalance
and label ambiguity in PLL. Nowadays, there are some researches have
been conducted on LT-PLL. Wang et al. (\citet{wang2018towards}) and
Liu et al. (\citet{liu2021partial}) addressed LT-PLL by employing
oversampling and regularization techniques. Solar (\citet{wang2022solar})
utilized optimal transport to mitigate pseudo-label bias by enforcing
class distribution priors. Additionally, RECORDS (\citet{hong2023long})
introduced a dynamic rebalancing strategy that adjusts logits based
on recovered class distributions. Furthermore, PLRIPL (\citet{xu2024pseudo})
presented a pseudo-label regularization technique by focusing on penalizing
pseudo labels of head classes. Meanwhile, HTC (\citet{jia2024long})
employed a dual-classifier model to handle samples from both head
and tail classes, leading to improved predictions across all class
distributions.
\begin{figure}
\centering{}\includegraphics[viewport=10bp 20bp 400bp 270bp,clip,scale=0.6]{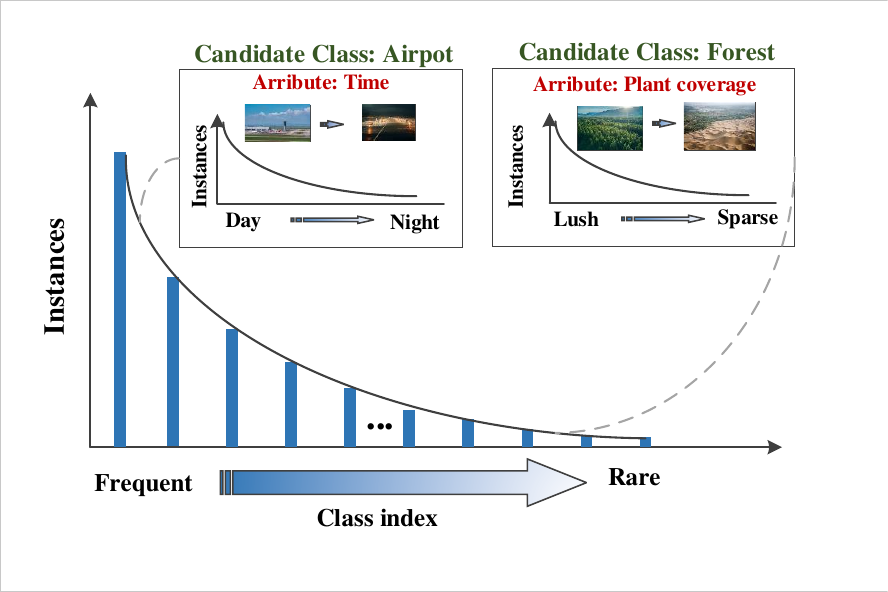}\caption{Practical example of a real-world long-tailed PLL dataset CIFAR100-LT.
There are both intra-class imbalanced samples for the majority candidate
labels ``Airport'' and the minority candidate labels ``Forest''.
Different from the existing methods that ignore the impact of intra-class
imbalance, our proposed GBRIP can consider both inter-class and intra-class
imbalance.}
\vspace{-1em}
\end{figure}

However, our study observed that existing IPLL and LT-PLL methods
rely on prior label confidence matrices (i.e., pseudo labels) derived
from inter-class features (i.e., inter-class differences). However,
these methods significantly ignore the impact of intra-class sample
imbalance (\citet{tang2022invariant}). For example, as shown in Figure
1, in the practical PLL CIFAR-100LT dataset, the majority candidate
class ``airport'' exhibits an inter-class feature imbalance phenomenon
of ``time'', such as fewer samples at night than during the day.
Similarly, the minority candidate class ``forest'' exhibits inter-class
imbalanced feature phenomena such as ``plant coverage''. Therefore, when
the intra-class feature space is not effectively mined and utilized,
such intra-class imbalance will lead to inaccurate and unreliable
label confidence matrices. Additionally, although some methods combine
sampling techniques to mitigate the impact of class imbalance, they
still struggle with the intra-class imbalance, as it's difficult for
them to sample and extract information from these intra-class imbalanced
samples effectively. Therefore, existing methods make it difficult
to solve the complexity of inter-class and intra-class imbalance in
IPLL, resulting in poor performance.

Motivated by the limitations of existing methods, we propose a novel
approach to imbalanced PLL called Granular Ball Representation for Imbalanced PLL (GBRIP). This method integrates two
key components: Coarse-grained Granular ball Representation (CGR)
and Multi-Center Loss (MCL). Given the challenges posed by PLL with
inexact supervision, CGR employs the superior GB technology
(\citet{xia2019granular,liuunlock,xia2022efficient,DBLP:journals/tkde/XiaLWGHS24,xie2024w,xie2024mgnr}) to significantly
unravel the intra-class and inter-class imbalance . Specially, by
using an unsupervised 2NN clustering method in CGR, the fine-grained
and imbalanced feature space is progressively divided into coarse-grained
GB representations with nearly equal size, where each sub GB acts
as a class. To address both inter-class and intra-class imbalances,
we construct a GB-based graph representation within this space using
a newly proposed weight measurement criterion. Unlike existing methods
that consider inter-class features only, our GB graph representation
captures both inter-class and intra-class imbalances, enabling the
construction of a more accurate label confidence matrix for guiding
label disambiguation in IPLL. During optimization, GBRIP employs a
loss function based on MCL, which focuses on the relationships between
samples and their respective centers within the granular balls. This
reduces the impact of outliers or hard-to-distinguish samples, leading
to more robust learning. The joint loss function in GBRIP effectively
mitigates the effects of inter-class and intra-class imbalances, ultimately
achieving superior performance.

\subsubsection{Contribution. }

We propose GBRIP, a novel method for IPLL that effectively addresses
both inter-class and intra-class imbalances. Using unsupervised clustering, GBRIP segments an imbalanced feature space into a balanced coarse-grained granular ball (GB) space. For this GB space, a novel weight measurement criterion is presented to enable the construction of a GB graph representation, effectively capturing the information of inter-class and intra-class imbalances, thus improving label confidence matrices and guiding disambiguation. GBRIP also introduces a Multi-Center Loss (MCL) function, optimizing a joint loss function. Extensive experiments validate GBRIP's superiority over state-of-the-art models, offering a robust solution for IPLL challenges.

\subsubsection{Compared to existing imbalanced PLL methods.}

The previously aforementioned IPLL and LT-PLL methods can solve the inter-class imbalanced PLL problems, while significantly
ignore the impact of intra-class imbalances (\citet{wang2022solar,jia2024long}), where such imbalanced
intra-class features can significantly degrade the accuracy and robustness
of label confidence construction. The proposed GBRIP can tackle the IPLL problem  more effectively by simultaneously considering
the impact of both inter-class and intra-class features on label confidence
using the newly developed GB graph representation space.

\subsubsection{Compared to existing GB methods.}

Existing GB-based methods can handle class imbalances (\citet{xia2021granular,xie2024three,DBLP:journals/tetci/XieZXZWWD24}), yet they significantly
depend on precise supervised class information, making them inapplicable
to PLL due to the lack of exact supervision. To tackle this issue,
we propose a novel GB-graph method that utilizes an unsupervised 2NN
clustering method and weight measurement to construct a coarse-grained
feature representation. This method effectively addresses sample imbalances
within and between sub-balls, resulting in a more accurate label confidence
matrix. Consequently, our approach successfully overcomes the challenges
of imbalanced PLL.

\section{Proposed Method}

\subsubsection{Problem setup.}

Let $\mathbf{X}\in R^{d}$ be the input space, and $\mathbf{Y}=\{y_{1},y_{2},\cdot\cdot\cdot,y_{L}\}$
be the label space with $L$ distinct categories. Given the PLL training
objects $\boldsymbol{D}=\{(\boldsymbol{x}_{i},\boldsymbol{S}_{i})\}_{i=1}^{N}$
where $N$ is the number of objects and $\boldsymbol{S}_{i}\subset\mathbf{Y}$
is the candidate label set for the sample $\boldsymbol{x}_{i}\in\mathbf{X}$
. We denote the $j$-th element of $\boldsymbol{S}_{i}$ as $\boldsymbol{S}_{i,j}$.
Here, $\boldsymbol{S}_{i,j}=1$ if the label $j$ is one of candidate
label for $\boldsymbol{x}_{i}$, and otherwise 0. The true label $y_{i}\in\boldsymbol{S}_{i}$
of $\boldsymbol{x}_{i}$ is concealed in $\boldsymbol{S}_{i}$. A
fundamental challenge in PLL is label disambiguation, i.e., identifying
the ground-truth label $y_{i}$ from the candidate label set $\boldsymbol{S}_{i}$.

The goal is to train a classifier $f:\mathbf{X}\rightarrow[0,1]^{L}$,
parameterized by $\theta$, that can perform predictions on unseen
testing data. Here, $f$ is the softmax output of a neural network,
and $f_{j}(\cdot)$ denotes the $j$-th entry. Let $\boldsymbol{P}=[\boldsymbol{p}_{1},...,\boldsymbol{p}_{N}]^{\top}=[p_{i,j}]_{N\times L}$
be the label confidence matrix. To perform label disambiguation, we
maintain a pseudo-label $\boldsymbol{p_{i}}$ for sample $\boldsymbol{x}_{i}$
where $p_{j}(\boldsymbol{x}_{i})$ donate the $j$-th entry. We train
the classifier with the cross-entropy loss $\mathcal{L}_{ce}(f;\boldsymbol{x}_{i},\boldsymbol{p}_{i})=\sum_{j=1}^{L}-p_{i,j}\log(f_{j}(\boldsymbol{x}_{i})).$

\subsubsection{Motivation of GBRIP.}

Our approach aims to maximize the distinction between imbalanced samples
from different classes while minimizing the distance between balanced
samples within the same class and making intra-class imbalanced samples
as far apart as possible. To address the challenges of imbalanced
PLL and inaccurate supervision, we propose a method that gradually
merges and separates samples in the feature space using unsupervised
clustering, creating a coarse-grained sample space. We introduce a
new GB-graph representation based on this coarse-grained space, which
leverages inter-class and intra-class sample information to construct
a more accurate label confidence matrix for effective disambiguation.
Figure 2 illustrates our method.
\begin{figure}
\centering{}\includegraphics[viewport=5bp 5bp 1355bp 1010bp,clip,scale=0.175]{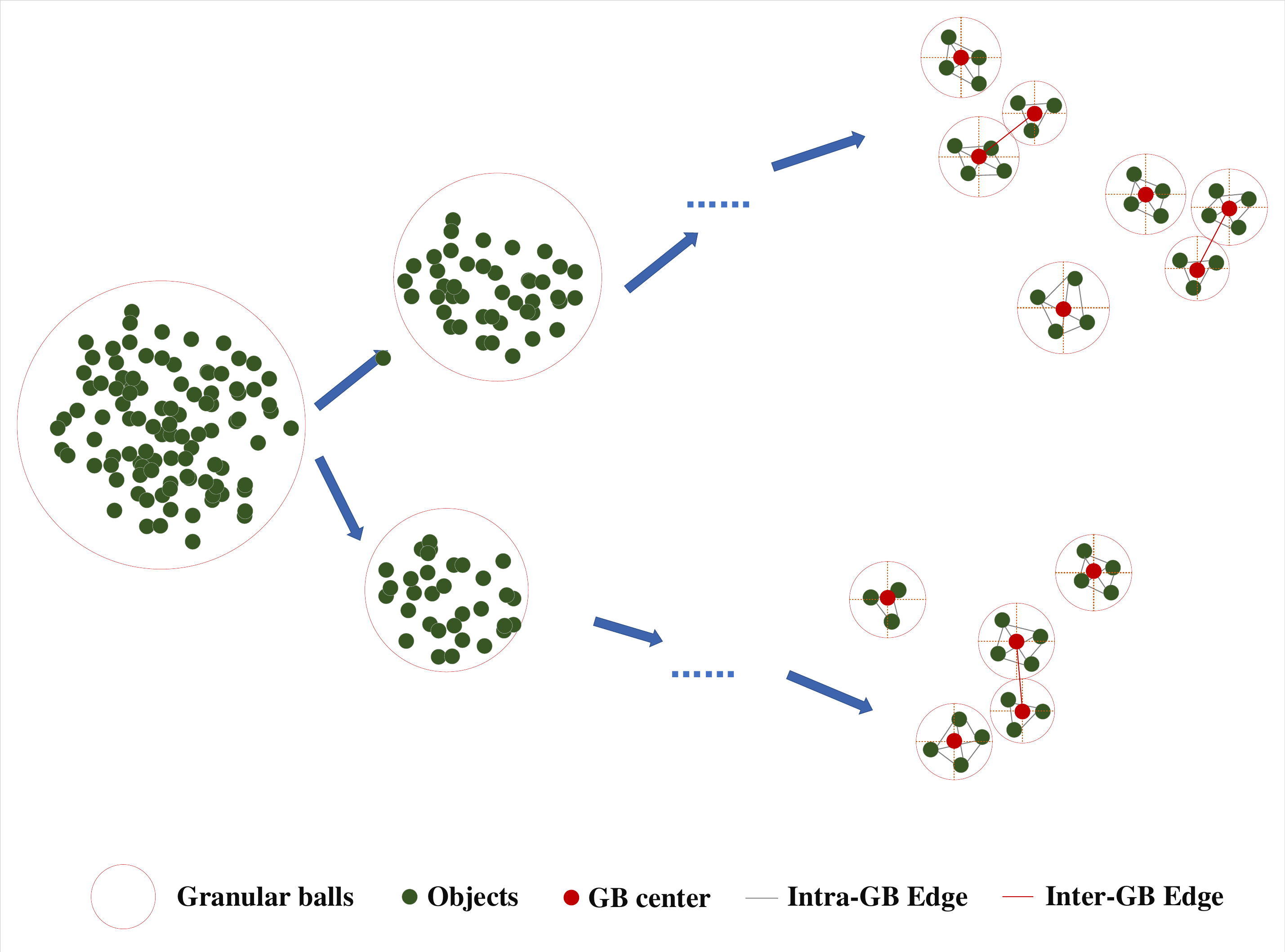}\caption{ The basic idea of GBRIP: Using 2NN unsupervised
clustering, imbalanced samples between and within classes are gradually
divided into smaller coarse-grained GB spaces, forming a new GB-graph
representation.}
\vspace{-1em}
\end{figure}

\subsubsection{The CGR module of GBRIP.}

Although GB can effectively cope with the challenge of imbalance,
they are limited by the challenge of PLL with inexact labeling, which
makes existing GB methods are difficult to construct accurate and
effective coarse-grained GB spaces. To this end, we propose a GB representation
method based on 2NN unsupervised clustering. First, for the feature
space $F(\mathbf{X})$ constructed by any feature extractor $\mathcal{F}(\cdot)$,
we regard it as a meta GB. Then, we use the 2NN method to cluster
$F(\mathbf{X})$, each time with 2 categories, and each category is
a sub-GB, which can separate samples with large differences from the
coarse-grained feature space. After that, the 2NN method is used cyclically,
which eventually separates the GBs with large inter-class differences
farther apart, while ensuring that the number of unbalanced samples
within each sub-GB is minimized.

The decision to further divide each GB depends on its size. Typically,
the maximum number of clusters is denoted as ``$n$'' chosen based
on established guidelines. If the current size of a GB exceeds the
defined limit ``$N$'', a 2-Means KNN clustering approach is employed
to split it into two smaller sub-balls. On the other hand, if the
size of the GB is less than or equal to ``$N$'', it is retained
as is. This method of division, based on the GB size, is designed
to strike an optimal balance between the granularity of the GBs and
the overall efficiency of the algorithm. By implementing this approach,
the algorithm effectively handles the trade-off between having too
many small clusters, which might reduce efficiency, and too few large
clusters, which might decrease the accuracy of classification or clustering
within the GBs.

Suppose $\text{\textbf{GB}}=\emptyset$ is an initialized GB space
for $F(\mathbf{X})$, and the initialized sub-GB $\boldsymbol{Q}=\boldsymbol{X}$,
 for each head GB $\boldsymbol{gb}_{i}\in  \boldsymbol{Q}$,
whether  $\boldsymbol{gb}_{i}$ needs to  be split should satisfy:
\begin{equation}
\begin{cases}
\boldsymbol{Q}=2NN(\boldsymbol{gb}_{i})\ add\ to\ tail, & \text{if}\ |\boldsymbol{gb}_{i}|>\sqrt{N},\\
\text{\textbf{GB}}=\text{\textbf{GB}}\cup\boldsymbol{gb}_{i}, & \text{otherwise}.
\end{cases}
\end{equation}

Accordingly, the generated GB representative space for $F(\mathbf{X})$
is denoted as:
\begin{equation}
\boldsymbol{\text{\textbf{GB}}=}\{\boldsymbol{gb}_{1},\boldsymbol{gb}_{2},...,\boldsymbol{gb}_{n}\}.
\end{equation}

For each $\boldsymbol{gb}_{i}\in\text{\textbf{GB}}$ with\emph{q}
objects $\boldsymbol{gb}_{i}=\{\boldsymbol{x}_{i}^{1},\boldsymbol{x}_{i}^{2},...,\boldsymbol{x}_{i}^{q}\}$
where $q$ is the number of object of $\boldsymbol{gb}_{i}$, the
center ($\boldsymbol{c_{i}}$), and the radius ($\boldsymbol{r_{i}}$)
of $\boldsymbol{gb}_{i}$ can be computed by:

\begin{equation}
\boldsymbol{c}_{i}=\frac{1}{\left|\boldsymbol{gb}_{i}\right|}\sum_{\boldsymbol{x}_{i}^{k}\in\boldsymbol{gb}_{i}}\boldsymbol{x}_{i}^{k},
\end{equation}
\begin{equation}
\boldsymbol{r}_{i}=\max\{dist(\boldsymbol{x}_{i}^{k},c_{i})\,|\,\boldsymbol{x}_{i}^{k}\in\boldsymbol{gb}_{i}\},
\end{equation}

where $|\boldsymbol{gb}_{i}|$ is the object number of the \emph{i}-th
GB, $\boldsymbol{x}_{i}^{k}\in\boldsymbol{gb}_{i}$ denotes the $k$-th
object of $i$-th ball, and $dist(\boldsymbol{x}_{i}^{k},c_{i})$
is the Euclidean distance. Specially, for each $\boldsymbol{gb}_{j}\in\text{\textbf{GB}}$,
$\boldsymbol{gb}_{i}\cap\boldsymbol{gb}_{j}=\emptyset$.

Based on $\boldsymbol{\text{\textbf{GB}}}$, a weighted graph
$\textbf{G}=(\boldsymbol{V},\boldsymbol{E})$ can be constructed where
$\boldsymbol{V}$ represents the training objects, and $\boldsymbol{E}$
is denoted as the similarity of two objects in $\boldsymbol{V}$,
i.e., the edges of graph $\textbf{G}$. It is worth noting that, different
from the traditional fully connected graph and \emph{K}-nearest neighbor
graph, the GB-based graph is constructed based on the given $\boldsymbol{\text{\textbf{GB}}}$
in this paper. Specifically, for each training object $\boldsymbol{x}_{i}^{k}\in\boldsymbol{gb}_{i}\subset\boldsymbol{GB}$,
let $\boldsymbol{B}(\boldsymbol{x}_{i}^{k})$ denote as all objects
in the $\boldsymbol{gb}_{i}$ who are in the same ball as $\boldsymbol{x}_{i}^{k}.$
For each $\boldsymbol{x}_{i}^{j}\in\boldsymbol{B}(\boldsymbol{x}_{i}^{k})$,
the edges $\boldsymbol{E}$ of GB-based graph between $\boldsymbol{x}_{i}^{k}$
and $\boldsymbol{x}_{i}^{j}$ can be denoted as:
\begin{equation}
\boldsymbol{E}=\{(\boldsymbol{x}_{i}^{k},\boldsymbol{x}_{i}^{j})\ |\ \boldsymbol{x}_{i}^{j}\in\boldsymbol{B}(\boldsymbol{x}_{i}^{k}),1\leq j\leq q\}.
\end{equation}

Specially, if $\boldsymbol{B}(\boldsymbol{x}_{i}^{k})=\emptyset$, $\boldsymbol{E}_{i}^{k}=\emptyset$. Notably, Eq. (5) only constructs
edges for samples within each GB, while neglecting the
edges of samples of inter-GBs. Since the difference between classes
and samples within classes is large, we construct edges for samples
between GBs according to the following criteria for two objects $\boldsymbol{x}_{i}^{k}\in\boldsymbol{gb_{i}}$
and $\boldsymbol{x}_{j}^{m}\in\boldsymbol{gb_{j}}$:
\begin{equation}
\begin{cases}
E(\boldsymbol{x}_{i}^{k},\boldsymbol{x}_{j}^{m})=\frac{1}{||\boldsymbol{c}_{i}+\boldsymbol{c}_{j}||}, & \text{if}\ ||\boldsymbol{c}_{i}+\boldsymbol{c}_{j}||<2\times\arg\max\{\boldsymbol{r}_{i},\boldsymbol{r}_{j}\},\\
E(\boldsymbol{x}_{i}^{k},\boldsymbol{x}_{j}^{m})=\emptyset, & \text{otherwise}.
\end{cases}
\end{equation}

Eq. (6) shows that if the distance between the centers of two balls
is less than twice the maximum radius of the two balls, we think that
the samples of the two balls are likely to be intra-class imbalanced
samples, then a connected graph of two GBs can be constructed from
the samples between them. On the contrary, the two GBs are regarded
as coarse-grained imbalanced samples between classes. Therefore, no
connected graph is established between the two GBs. In other words,
the weight of imbalanced samples between the two classes in the GB
graph is 0.

For the given set of edges $\boldsymbol{E}$, a weighted matrix is
defined as $\textbf{W}=[\boldsymbol{w}_{i,j}]_{N\times N}$ where
$\boldsymbol{w}_{i,j}>0$ if $(\boldsymbol{x}_{i},\boldsymbol{x}_{j})\in E$,
and $\boldsymbol{w}_{i,j}=0$, otherwise. In this paper, we hope to
reconstruct $\boldsymbol{x}_{i}^{k}$ through the weighted sum of
adjacent samples in the $\text{\textbf{GB}}$ space of object $\boldsymbol{x}_{i}^{k}$.
Let $\boldsymbol{w}$ be the weight vector of \textbf{$\boldsymbol{x}_{i}^{k}$}
and its adjacent samples in $\text{\textbf{GB}}$, and the weight $\boldsymbol{w}_{i}^{k,j}$
between $\boldsymbol{x}_{i}^{k}$ and its adjacent object $\boldsymbol{x}_{i}^{j}$ can be obtained by solving the
following optimization problem:
\begin{equation}
\min_{\boldsymbol{w}_{i}^{k}}\left\Vert \boldsymbol{x}_{i}^{k}-\sum_{j=1}^{q-1}\boldsymbol{w}_{i}^{k,j}\cdot\boldsymbol{x}_{i}^{j}\right\Vert ^{2},
\end{equation}
\[
s.t.\ \boldsymbol{w}_{i}^{k,j}\geq0,\ \forall\boldsymbol{x}_{i}^{j}\in\boldsymbol{B}(\boldsymbol{x}_{i}^{k}).
\]

The optimization process of Eq. (7) aims to minimize the reconstruction
error of the nearest neighbors of $\boldsymbol{x}_{i}^{k}$. Therefore,
it is necessary to assign higher weights to objects that make a significant
contribution to the reconstruction so that they can have a more significant
impact on the iterative label transfer process. To achieve this, we
need to solve a non-negative linear least squares problem to determine
the weight vector $\textbf{W}$ that quantifies the relationship between
each object and its nearest neighbor. Any quadratic programming solver
can be used to find the optimal solution to this optimization problem.
The proposed approach ensures that the neighbors that contribute the
most in terms of reconstruction accuracy are prioritized, resulting
in an overall improvement in the effectiveness of the label disambiguation.

\subsubsection{Label Disambiguation of GBRIP by Multi Center Loss (MCL). }

To obtain the disambiguation label sets, for the label propagation
confidence matrix $\textbf{P}=[\boldsymbol{P}_{1},\boldsymbol{P}_{2},...,\boldsymbol{P}_{n}]_{n\times L}$,
the label confidence vector $\boldsymbol{P}_{i}^{k}=[p_{i,j}^{k}]_{1\times L}^{\top}$
for object $\boldsymbol{x}_{i}^{k}\in\boldsymbol{gb}_{i}\subset\text{\textbf{GB}}$
is initialized as:
\begin{equation}
p_{i,j}^{k}=\begin{cases}
\frac{\boldsymbol{S}_{i,j}^{k}f_{j}(\boldsymbol{x}_{i}^{k})}{\boldsymbol{W}_{i}^{k}\times\sum_{j=1}^{L}\boldsymbol{S}_{i,j}^{k}f_{j}(\boldsymbol{x}_{i}^{k})}, & \text{if}\ y_{j}\in\boldsymbol{S}_{i},\\
0, & \text{otherwise},
\end{cases}
\end{equation}

where $p_{i,j}^{k}\in[0,1]$ is denoted the label confidence of label
$y_{j}$ as the ground-truth label of object $\boldsymbol{x}_{i}$,
and $\boldsymbol{W}_{i}^{k}=\sqrt{\sum_{m=1}^{n}\boldsymbol{w}_{i}^{k,m}}$
can be obtained by Eq. (7).

We have developed a CGR-based feature space to represent a well-balanced
feature space to a certain extent. It is crucial to guarantee that
the center of each GB can accurately represent the respective samples
of the GB. To accomplish this, we have imposed constraints on the
CGR feature space to improve the representativeness of the GB centers.
While the segmented balls address intra-class imbalances, further
refinement is needed to handle abnormal or highly unbalanced samples
within the balls that are challenging to differentiate. To address
this, we have devised a multi-center loss function that emphasizes
the GB centers and the balls themselves, ensuring that the representation
remains robust and precise:

\begin{equation}
\min_{\theta,p}\sum_{i=1}^{n}\sum_{k=1}^{i}\mathcal{L}_{cls}\left (F(x_{i}^{k}),\boldsymbol{P}_{i}^{k};\theta\right),
\end{equation}
\begin{align}
s.t.\quad\theta\in\arg\min_{\Theta} & \sum_{i=1}^{n}\sum_{k=1}^{i}\left\Vert F(x_{i}^{k})-\boldsymbol{c}_{i}\right\Vert _{2}.\nonumber
\end{align}

In Eq. (9), the component $\mathcal{L}_{cls}$ denotes the classifier
loss during the learning phase, and the standard cross-entropy loss
($\mathcal{L}_{ce}$) is used in this paper. The second component
represents the center loss, aiming to guarantee the accurate representation
of each ball's center. This approach effectively addresses inter-class
and intra-class imbalances, promoting balanced and resilient learning
outcomes:
\begin{align}
\min_{\theta,p}\sum_{i=1}^{n}\sum_{k=1}^{i}\mathcal{L}= & \min_{\theta,p}\sum_{i=1}^{n}\sum_{k=1}^{i}\mathcal{L}_{ce}+\mathcal{L}_{mc}\nonumber \\
= & \min_{\theta,p}\sum_{i=1}^{n}\sum_{k=1}^{i}\sum_{j=1}^{L}-p_{i,j}^{k}\log(f_{j}(\boldsymbol{x}_{i}^{k}))\\
 & +\lambda\cdot\left\Vert F(\boldsymbol{x}_{i}^{k})-\boldsymbol{c}_{i}\right\Vert _{2}.\nonumber
\end{align}

Therefore, the final objective function of our proposed GBRIP is denoted
as:
\begin{align}
\mathcal{L}_{GBRIP}= & \lambda_{1}\mathcal{L}_{ce}+\lambda_{2}\mathcal{L}_{mc}+\lambda_{3}\mathcal{L}_{pr}\nonumber \\
= & \min_{\theta,p}\sum_{i=1}^{n}\sum_{k=1}^{i}\sum_{j=1}^{L}\left[\lambda_{1}\left(-p_{i,j}^{k}\log(f_{j}(\boldsymbol{x}_{i}^{k}))\right)\right.\\
 & +\lambda_{2}\times\left\Vert F(\boldsymbol{x}_{i}^{k})-\boldsymbol{c}_{i}\right\Vert _{2}\nonumber \\
 & +\left.\lambda_{3}\times p_{i,j}^{k}\log r_{j}\right],\nonumber
\end{align}

where the third term $\mathcal{L}_{pr}$ is the regularization item
of the category, which will keep the pseudo labels away from the prior
distribution $r$. Similar to previous work (\citet{jia2024long,xu2024pseudo}),
we adopt the method of alternately optimizing $w$ and $j$. For constraint
$p_{i,j}^{k}=0$ for $\boldsymbol{S}_{i,j}^{k}=0$, we can delete
the items related to $p_{i,j}^{k}$ in the optimization goal if $\boldsymbol{S}_{i,j}^{k}=0$.
By using Lagrange multiplier method, $p$ can be optimized by:

\begin{equation}
p_{i,j}^{k}=\frac{\boldsymbol{S}_{i,j}^{k}f_{i,j}\boldsymbol{u}_{j}^{-\lambda_{3}}}{\boldsymbol{W}_{i}^{k}\times\sum_{j=1}^{L}\boldsymbol{S}_{i,j}^{k}f_{i,j}^{\lambda}\boldsymbol{u}_{j}^{-\lambda_{3}}}.
\end{equation}

One of the key difference between long-tail learning and long-tail
PLL in that the number of each category is unknown, which requires
us to estimate $\boldsymbol{u}$ based on the information of training data. To
estimated $\boldsymbol{u}$, we initialize $\boldsymbol{u}$ to be uniformly distributed $[1/c,...,1/c]$,
and update it by using a moving-average strategy to ensure the stability
of updating:
\begin{equation}
\boldsymbol{u}\leftarrow\theta \boldsymbol{u}+(1-\theta)\frac{1}{n}\sum_{i=1}^{n}\sum_{k=1}^{q}\prod(j=\arg\max_{j^{'}\in\boldsymbol{S}_{i}^{k}}f_{j^{'}}(\boldsymbol{x}_{i}^{k})),
\end{equation}

where $\theta\in[0,1]$ is a preset scalar. One advantage of this estimation
method is that assuming our classifier can fully predict accurately,
the estimated $\boldsymbol{u}$ can approach the true $\boldsymbol{u}$.

\section{Experiments}

\subsection{Experimental Settings}

\subsubsection{Datasets.}

We evaluated our method on two long-tailed datasets: CIFAR10-LT and
CIFAR100-LT. The training images were randomly removed class-wise
to create a predefined imbalance ratio $\gamma=\frac{n_{j}}{n_{L}}$,
where $n_{j}$ represents the number of images in the $j$-th class.
For convenience, class indices were sorted by sample size in descending
order, ensuring $n1\ge n2\ge...\ge n_{L}$ with a consistent ratio
between consecutive classes. To generate partially labeled datasets,
we manually flipped negative labels $(\hat{y}\neq y$) to false-positive
labels with a probability $\psi=P(\hat{y}\in Y|\hat{y}\neq y)$, following
previous works (\citet{jia2024long,xu2024pseudo}). The final candidate
label set included the ground-truth label and flipped false-positive
labels. We selected $\gamma=\{50,100,200\},\psi\in\{0.3,0.5\}$ for
CIFAR10-LT and $\gamma=\{10,20,50\},\psi\in\{0.05,0.1\}$ for CIFAR100-LT.
We report the mean and standard deviation from three independent runs
with the same random seed for all experiments, selecting the model
with the best validation performance as the final model.
\begin{table*}
\caption{Accuracy comparisons on CIFAR10-LT and CIFAR100-LT under various flipping
probability $\psi$ and imbalance ratio $\gamma$. The best results
are marked in bold and the second-best marked in Italic.}

\centering{}%
\begin{tabular}{c|c|c|c|c|c|c}
\hline
\multirow{3}{*}{Methods} & \multicolumn{6}{c}{CIFAR10-LT}\tabularnewline
\cline{2-7} \cline{3-7} \cline{4-7} \cline{5-7} \cline{6-7} \cline{7-7}
 & \multicolumn{3}{c|}{$\psi=0.3$} & \multicolumn{3}{c}{$\psi=0.5$}\tabularnewline
\cline{2-7} \cline{3-7} \cline{4-7} \cline{5-7} \cline{6-7} \cline{7-7}
 & $\gamma=50$ & $\gamma=100$ & $\gamma=200$ & $\gamma=50$ & $\gamma=100$ & $\gamma=200$\tabularnewline
\hline
MSE & 61.13\textpm 1.08 & 52.59\textpm 0.48 & 48.09\textpm 0.45 & 49.61\textpm 1.42 & 43.90\textpm 0.77 & 39.52\textpm 0.70\tabularnewline
VALEN & 58.34\textpm 1.05 & 50.20\textpm 6.55 & 46.98\textpm 1.24 & 40.04\textpm 1.80 & 37.10\textpm 0.88 & 36.61\textpm 0.57\tabularnewline
LWS & 44.51\textpm 0.03 & 43.60\textpm 0.12 & 42.33\textpm 0.58 & 24.62\textpm 9.67 & 27.33\textpm 1.84 & 28.74\textpm 1.86\tabularnewline
PRODEN & 81.95\textpm 0.19 & 71.09\textpm 0.54 & 63.00\textpm 0.54 & 66.00\textpm 3.60 & 62.17\textpm 3.36 & 54.65\textpm 1.00\tabularnewline
PiCO & 75.42\textpm 0.49 & 67.73\textpm 0.64 & 61.12\textpm 0.67 & 72.33\textpm 0.08 & 63.25\textpm 0.64 & 53.92\textpm 1.64\tabularnewline
\hline
RECORDS & 84.57\textpm 0.36 & 77.95\textpm 0.36 & 71.67\textpm 0.57 & 80.28\textpm 1.11 & 74.05\textpm 1.11 & 63.75\textpm 0.47\tabularnewline
Solar & 83.80\textpm 0.52 & 76.64\textpm 1.66 & 67.47\textpm 1.05 & 81.38\textpm 2.84 & 74.16\textpm 3.03 & 62.12\textpm 1.64\tabularnewline
PLRIPL & 87.25\textpm 0.51 & 81.74\textpm 0.53 & 74.07\textpm 1.45 & 85.86\textpm 1.01 & 78.38\textpm 0.37 & 65.76\textpm 2.86\tabularnewline
HTC & \emph{88.14\textpm 0.94} & \emph{85.66\textpm 1.44} & \emph{80.57\textpm 1.40} & \emph{86.11\textpm 1.07} & \emph{83.25\textpm 2.24} & \emph{77.71\textpm 1.12}\tabularnewline
GBRIP & \textbf{91.54\textpm 0.12} & \textbf{88.97\textpm 1.55} & \textbf{85.78\textpm 1.55} & \textbf{89.79\textpm 1.66} & \textbf{86.71\textpm 0.11} & \textbf{83.97\textpm 1.66}\tabularnewline
\hline
\hline
\multirow{3}{*}{Methods} & \multicolumn{6}{c}{CIFAR100-LT}\tabularnewline
\cline{2-7} \cline{3-7} \cline{4-7} \cline{5-7} \cline{6-7} \cline{7-7}
 & \multicolumn{3}{c|}{$\psi=0.05$} & \multicolumn{3}{c}{$\psi=0.1$}\tabularnewline
\cline{2-7} \cline{3-7} \cline{4-7} \cline{5-7} \cline{6-7} \cline{7-7}
 & $\gamma=10$ & $\gamma=20$ & $\gamma=50$ & $\gamma=10$ & $\gamma=20$ & $\gamma=50$\tabularnewline
\hline
MSE & 49.92\textpm 0.64 & 43.94\textpm 0.86 & 37.74\textpm 0.40 & 42.99\textpm 0.47 & 37.19\textpm 0.72 & 31.49\textpm 0.35\tabularnewline
VALEN & 49.12\textpm 0.58 & 42.05\textpm 1.52 & 35.62\textpm 0.43 & 33.39\textpm 0.65 & 30.37\textpm 0.11 & 24.93\textpm 0.87\tabularnewline
LWS & 48.85\textpm 2.16 & 35.88\textpm 1.29 & 19.22\textpm 8.56 & 6.10\textpm 2.05 & 7.16\textpm 2.03 & 5.15\textpm 0.36\tabularnewline
PRODEN & 60.36\textpm 0.52 & 54.33\textpm 0.21 & 45.83\textpm 0.31 & 57.91\textpm 0.41 & 51.09\textpm 0.48 & 41.74\textpm 0.41\tabularnewline
PiCO & 54.05\textpm 0.37 & 46.93\textpm 0.65 & 38.74\textpm 0.11 & 46.49\textpm 0.46 & 39.80\textpm 0.34 & 34.97\textpm 0.09\tabularnewline
\hline
RECORDS & 63.21\textpm 0.17 & 57.60\textpm 1.99 & 49.04\textpm 1.57 & 60.52\textpm 1.77 & 54.73\textpm 0.80 & 45.47\textpm 0.74\tabularnewline
Solar & 64.75\textpm 0.07 & 56.47\textpm 0.76 & 46.18\textpm 0.85 & 61.82\textpm 0.71 & 53.03\textpm 0.56 & 40.96\textpm 1.01\tabularnewline
PLRIPL & \emph{65.83\textpm 0.43} & 58.62\textpm 0.61 & 48.73\textpm 0.25 & \emph{63.89\textpm 0.63} & 54.49\textpm 0.64 & 45.74\textpm 0.70\tabularnewline
HTC & 64.59\textpm 0.74 & \emph{61.13\textpm 1.15} & \emph{53.26\textpm 1.90} & 62.77\textpm 0.45 & \emph{60.53\textpm 1.45} & \emph{51.26\textpm 1.31}\tabularnewline
GBRIP & \textbf{67.15\textpm 1.24} & \textbf{63.42\textpm 0.97} & \textbf{57.94\textpm 0.38} & \textbf{64.22\textpm 1.55} & \textbf{62.37\textpm 1.89} & \textbf{56.91\textpm 0.58}\tabularnewline
\hline
\end{tabular}
\end{table*}

\subsubsection{Baselines.}

We compared our method with nine state-of-the-art PLL methods, categorizing
them based on whether they consider class imbalance. The methods that
address imbalance or long-tailed PLL include four SOTA methods such
as RECORDS (\citet{hong2023long}), Solar (\citet{wang2022solar}),
PLRIPL (\citet{xu2024pseudo}), and HTC (\citet{jia2024long}). Additionally,
we compared our approach with five recent SOTA PLL methods that do
not consider imbalance, including MSE (\citet{feng2020learning}),
VALEN (\citet{xu2021instance}) LWS \citet{wen2021leveraged}, PRODEN,
and PiCO (\citet{wang2022pico}). All hyper-parameters were tuned
according to the settings provided in the original papers.

\subsubsection{Implementation details.}

We utilized an 18-layer ResNet as the feature backbone for our experiments.
The model was trained for 1000 epochs using the standard SGD optimizer
with a momentum of 0.9. The initial learning rate was set to 0.01
and decayed using a cosine learning rate schedule. The batch size
was fixed at 256. These configurations were applied consistently across
our method and all baseline models to ensure a fair comparison. We
conducted a pre-estimation training phase to obtain coarse-grained
class priors, running the model 100 times on CIFAR10-LT and 20 times
on CIFAR100-LT, following the approach used in previous work. After
this phase, the model weights were initialized, and training resumed
with the obtained class priors. For our method, GBRIP, the hyper-parameters
were set as follows: $\lambda_{1}=0.5$, $\lambda_{2}=0.5$ and $\lambda_{3}=0.1$.
The moving average parameter $\mu$ for the class prior estimate was
set to 0.1/0.05 in the first phase and fixed at 0.01 afterward. For
class-reliable sample selection, the parameter $\rho$ was linearly
increased from 0.2 to 0.5/0.6 over the first 50 epochs. We incorporated
consistency loss and mixture into all baseline models except PiCO
for a fair comparison. The mixture coefficients were sampled from
a $\beta(4,4)$ distribution. All experiments were conducted three
times with different random seeds, and we reported the mean and standard
deviation of the results.

\subsection{Experimental Results}

\subsubsection{GBRIP achieves optimal results. }

As presented in Table 1, GBRIP outperforms the state-of-the-art (SOTA) methods
on CIFAR10-LT and CIFAR100-LT datasets across various $\psi$ and
$\gamma$. The findings indicate that GBRIP significantly outperforms
all competing methods substantially. Specifically, on the CIFAR10-LT
dataset with $\psi=0.3$ and an imbalance ratio $\gamma=200$, GBRIP
exhibits a 5.21\% improvement over the optimal baseline, highlighting
its effectiveness in handling scenarios with high imbalance. Furthermore,
GBRIP shows its superiority under $\psi=0.5$ and $\gamma=200$, achieving
a 6.26\% performance gain over the best-performing baseline method.
GBRIP consistently outperforms other methods on the CIFAR100-LT dataset,
which is more challenging due to its larger number of classes and
stronger label ambiguity. Notably, with $\psi=0.1$ and $\gamma=50$,
GBRIP surpasses the best baseline by 5.65\%. Even as the imbalance
ratio increases, GBRIP maintains its competitive edge, outperforming
other methods significantly across all settings. These observations
validate the superiority and effectiveness of GBRIP, particularly in
scenarios with high imbalance and label ambiguity.

\subsubsection{Results on different groups of labels. }

The findings presented in Table 2 demonstrate that GBRIP achieves optimal performance across various class distributions in both CIFAR10-LT
($\psi=0.5,\gamma=100$) and CIFAR100-LT ($\psi=0.1,\gamma=20$).
The dataset is categorized into Many, Medium, and Few groups based
on the number of samples per class. GBRIP consistently outperforms
all competing methods in overall accuracy and within each class distribution
category. On CIFAR10-LT, GBRIP achieves an exceptional overall accuracy
of 86.71\%, with strong performance across all groups: 97.69\% for
Many-shot classes, 85.12\% for Medium-shot classes, and 74.23\% for
Few-shot classes. Notably, GBRIP surpasses the second-best method,
HTC, by 3.46\% overall accuracy and 3.10\% and 3.07\% in Medium-shot
and Few-shot classes, respectively. This demonstrates GBRIP's ability
to maintain high accuracy even in classes with fewer samples. Similarly,
on CIFAR100-LT, GBRIP showcases the outstanding results, with an overall
accuracy of 62.37\%. GBRIP performs exceptionally well across all class
distributions, achieving 80.49\% for Many-shot classes, 66.22\% for
Medium-shot classes, and 52.10\% for Few-shot classes, and
outperforms HTC by 1.84\% overall accuracy and shows significant improvements
in Medium-shot and Few-shot classes, with margins of 5.01\% and 5.89\%,
respectively. The consistent superiority across different class distributions
validate GBRIP's effectiveness in addressing IPLL.
\begin{table*}
\caption{Different shots accuracy comparisons on CIFAR10-LT ($\psi=0.5,\gamma=100$)
and CIFAR100- LT $(\psi=0.1,\gamma=20$). The best results are marked
in bold and the second-best marked in italic.}

\centering{}%
\begin{tabular}{c|c|c|c|c|c|c|c|c}
\hline
\multirow{2}{*}{Methods} & \multicolumn{4}{c|}{CIFAR10-LT} & \multicolumn{4}{c}{CIFAR100-LT}\tabularnewline
\cline{2-9} \cline{3-9} \cline{4-9} \cline{5-9} \cline{6-9} \cline{7-9} \cline{8-9} \cline{9-9}
 & All & Many  & Medium  & Few  & Total & Many  & Medium  & Few\tabularnewline
\hline
MSE  & 43.90  & 81.11  & 42.03  & 9.18  & 37.19 & 57.46  & 37.57  & 16.53\tabularnewline
LWS  & 27.33  & 89.09  & 1.52  & 0.00  & 7.16  & 20.80  & 0.86  & 0.00\tabularnewline
VALEN  & 37.10  & 85.30  & 28.78  & 0.00  & 30.37  & 58.74  & 16.25  & 0.07\tabularnewline
CAVL & 37.51 & 82.67 & 16.43 & 0.00 & 18.29 & 48.12 & 2.57 & 0.00\tabularnewline
PRODEN  & 62.17  & 96.83  & 72.18  & 14.17  & 51.09  & 76.86  & 43.14  & 5.43\tabularnewline
PiCO  & 63.25  & 93.33  & 66.14  & 29.30  & 39.80 & 70.75  & 42.42  & 6.14\tabularnewline
\hline
RECORDS & 67.74 & 88.17 & 71.10 & 35.00 & 54.73 & 76.09 & 45.65 & 8.52\tabularnewline
Solar  & \emph{83.80 } & \emph{96.50 } & 76.01  & 49.34  & 53.03  & 74.33  & 54.09  & 30.62\tabularnewline
PLRIPL  & 78.38 & 85.11 & 78.75 & 71.16 & 54.49  & 76.06 & 60.33 & 40.17\tabularnewline
HTC & 83.25 & 91.58 & \emph{80.83} & \emph{71.13} & \emph{60.53} & \emph{78.52} & \emph{61.21} & \emph{46.21}\tabularnewline
GBRIP  & \textbf{86.71} & \textbf{97.69} & \textbf{85.12} & \textbf{74.23} & \textbf{62.37} & \textbf{80.49} & \textbf{66.22} & \textbf{52.10}\tabularnewline
\hline
\end{tabular}
\end{table*}

\vspace{-1em}
\subsubsection{Results on real-world PLL datasets. }

This section evaluates GBRIP's performance on four classical real-world
PLL datasets: Lost, Bird Song(Bird. S), Soccer Player(Soccer. P),
and Yahoo! News (Yahoo. N). These datasets are naturally imbalanced,
highlighting the challenges addressed by GBRIP. As shown in Table 3,
GBRIP consistently outperforms all the compared methods across these
diverse datasets. On the Lost dataset, GBRIP achieves an accuracy of
85.73\%, surpassing the second-best method, HTC, by a margin of 3.02\%.
This noteworthy improvement demonstrates GBRIP's capability to handle
imbalanced real-world data effectively. Similarly, on the Bird Song
dataset, GBRIP achieves the highest accuracy of 78.93\%, outperforming
HTC by 0.95\%. The Soccer Player dataset, known for its severe imbalance
with an extremely high imbalance ratio, presents a particularly challenging
scenario. Despite this, GBRIP achieves an accuracy of 62.78\%, which
is 2.67\% higher than the second-best method, HTC, indicating its
robustness in extremely imbalanced conditions. On the Yahoo! News
dataset, GBRIP also demonstrates competitive performance with an accuracy
of 72.38\%, closely following HTC, which achieves 72.61\%. Although
the difference is marginal, GBRIP still exhibits strong performance
across all datasets. These results highlight GBRIP's superiority in
addressing the imbalanced PLL problem in real-world datasets.GBRIP consistently outperforms other methods across various datasets, including those with severe imbalances, underscoring its effectiveness and robustness in practical applications.
\begin{table}
\caption{Performance comparisons on four real-world partial-label learning
datasets}

\centering{}%
\begin{tabular}{c|cccc}
\hline
\multirow{1}{*}{Methods} & Lost & Bird. S & Soccer. P & Yahoo. N\tabularnewline
\hline
VALEN & 74.11 & 71.59 & 57.16 & 67.93\tabularnewline
PRODEN & 78.98 & 71.81 & 57.12 & 67.87\tabularnewline
RECORDS & 77.58 & 73.86 & 58.54 & 69.13\tabularnewline
Solar & 77.86 & 72.05 & 57.94 & 67.62\tabularnewline
PLRIPL & 79.05 & 75.22 & 58.29 & 69.77\tabularnewline
HTC & 82.71 & 77.98 & 60.11 & \textbf{72.61}\tabularnewline
GBRIP & \textbf{85.73} & \textbf{78.93} & \textbf{62.78} & 72.38\tabularnewline
\hline
\end{tabular}
\end{table}

\subsubsection{Results on real-world long-tailed learning data. }

To verify the effectiveness of GBRIP on real-world imbalanced datasets,
we conducted experiments on the large-scale SUN397 dataset, which
contains 108,754 RGB images across 397 scene classes. We set the batch
size to 128 for these experiments and held out 50 samples per class
for testing. This setup resulted in a training set with an imbalanced
ratio of approximately 46 (2311/50). We trained the model for 20 epochs
for distribution estimation and 200 epochs for regular training. Additionally,
we synthesized a partial-label dataset with an ambiguity degree $\psi=0.1$
of to further test GBRIP's performance under real-world conditions.
As shown in Fig. 3, GBRIP significantly outperforms the baselines across
all categories, including Many-shots, Medium-shots, and Few-shots.
This consistent superiority highlights GBRIP's robustness and effectiveness
in handling highly imbalanced and complex datasets like SUN397. The
results further validate GBRIP's capability to maintain high performance
even in challenging real-world scenarios.
\begin{figure}[H]
\centering{}\includegraphics[scale=0.65]{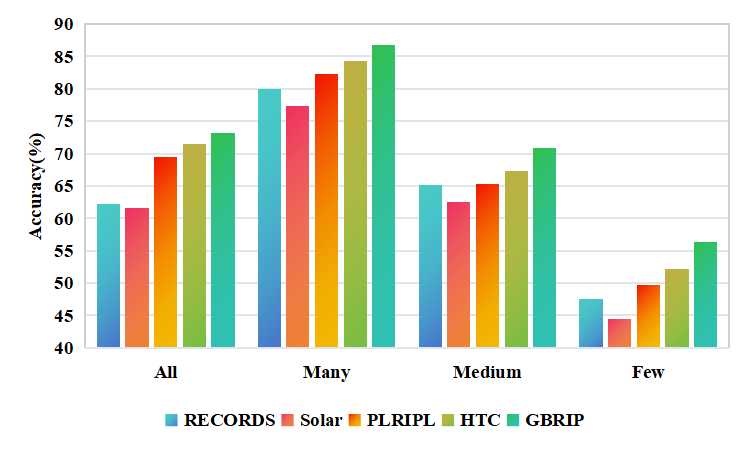}\caption{Performance comparisons on the SUN397 dataset with $\psi=0.05$.}
\vspace{-1em}
\end{figure}

\subsubsection{Ablation studies of all the components contribute to GBRIP.}

We conducted ablation studies to evaluate the effectiveness of each
component in GBRIP. Removing the CGR and MCL modules (w/o C+M) resulted
in a significant performance drop, confirming their critical roles.
The absence of MCL alone (w/o M) or CGR (replaced with KMeans, w/o
C({*}K)) also led to notable accuracy declines, highlighting their
importance. Removing Mixups (w/o U) further reduced performance, demonstrating
their contributions to robustness and training effectiveness. The
removal of Logit adjustment (w/o L) and sample selection (w/o S) also
negatively impacted results, underscoring their necessity for addressing
class imbalance and ensuring model reliability. Overall, each component
of GBRIP significantly contributes to its superior performance on imbalanced
and partial-label datasets.

\begin{table}[H]
\caption{Ablation study of our method on LT-PLL datasets CIFAR10 $\psi=0.5,\gamma=100$)}

\centering{}%
\begin{tabular}{c|c|c|c|c}
\hline
\multirow{1}{*}{Methods} & All & Many & Medium & Few\tabularnewline
\hline
GBRIP & \textbf{86.71} & \textbf{97.69} & \textbf{85.12} & \textbf{74.23}\tabularnewline
\hline
GBRIP w/o C+M & 37.06 & 67.32 & 30.73 & 1.10\tabularnewline
GBRIP w/o M & 70.59 & 73.15 & 75.47 & 57.63\tabularnewline
GBRIP w/o C({*}K) & 68.40 & 87.50 & 69.58 & 46.19\tabularnewline
\hline
GBRIP w/o U & 73.60 & 90.98 & 70.23 & 51.69\tabularnewline
GBRIP w/o L & 75.07 & 81.58 & 76.95 & 59.03\tabularnewline
GBRIP w/o S & 75.53 & 82.82 & 75.76 & 60.93\tabularnewline
\hline
\end{tabular}
\end{table}

\section{Conclusion}

In our study, we have introduced GBRIP, which is specifically designed to address the challenges of imbalanced partial-label learning (IPLL), focusing on tackling inter-class and intra-class imbalances. GBRIP utilizes a coarse-grained granular ball (GB) based feature representation to transform an imbalanced fine-grained feature space into a balanced one. This transformation is achieved through an unsupervised 2NN clustering criterion that segments the feature space into approximately equal-sized sub-GBs. Within this GB space, we have developed a new weight measurement criterion to create a GB graph representation, capturing inter-class and intra-class imbalances. This enables the accurate construction of a label confidence matrix, guiding effective label disambiguation. Additionally, GBRIP incorporates a multi-center loss (MCL) function that optimizes a joint loss function, reducing outliers' impact and enhancing learning robustness. Experimental results across benchmark datasets have validated GBRIP's superiority over existing PLL methods, demonstrating its effectiveness in handling imbalanced PLL scenarios.

\section{Acknowledgments}
This work was supported in part by the NSFC/Research Grants
Council (RGC) Joint Research Scheme under grant:N\_HKBU214/21, the General Research Fund of RGC under the grants: 12202622, 12201323, and the RGC Senior Research Fellow Scheme under the grant: SRFS2324-2S02; in part by the National Natural Science Foundation of China (No.62366019), and the Jiangxi Provincial Natural Science Foundation, China (No.20242BAB23014).
\bibliography{aaai25}

\begin{thebibliography}{36}
\providecommand{\natexlab}[1]{#1}

\bibitem[{Chen, Patel, and Chellappa(2017)}]{chen2017learning}
Chen, C.-H.; Patel, V.~M.; and Chellappa, R. 2017.
\newblock Learning from ambiguously labeled face images.
\newblock \emph{IEEE transactions on pattern analysis and machine
  intelligence}, 40(7): 1653--1667.

\bibitem[{Cour, Sapp, and Taskar(2011)}]{cour2011learning}
Cour, T.; Sapp, B.; and Taskar, B. 2011.
\newblock Learning from partial labels.
\newblock \emph{The Journal of Machine Learning Research}, 12: 1501--1536.

\bibitem[{Feng et~al.(2020)Feng, Kaneko, Han, Niu, An, and
  Sugiyama}]{feng2020learning}
Feng, L.; Kaneko, T.; Han, B.; Niu, G.; An, B.; and Sugiyama, M. 2020.
\newblock Learning with multiple complementary labels.
\newblock In \emph{International Conference on Machine Learning}, 3072--3081.
  PMLR.

\bibitem[{Hong et~al.(2023)Hong, Yao, Zhou, Zhang, and Wang}]{hong2023long}
Hong, F.; Yao, J.; Zhou, Z.; Zhang, Y.; and Wang, Y. 2023.
\newblock Long-tailed partial label learning via dynamic rebalancing.
\newblock \emph{arXiv preprint arXiv:2302.05080}.

\bibitem[{H{\"u}llermeier and Beringer(2006)}]{hullermeier2006learning}
H{\"u}llermeier, E.; and Beringer, J. 2006.
\newblock Learning from ambiguously labeled examples.
\newblock \emph{Intelligent Data Analysis}, 10(5): 419--439.

\bibitem[{Jia et~al.(2024)Jia, Peng, Wang, and Zhang}]{jia2024long}
Jia, Y.; Peng, X.; Wang, R.; and Zhang, M.-L. 2024.
\newblock Long-Tailed Partial Label Learning by Head Classifier and Tail
  Classifier Cooperation.
\newblock In \emph{Proceedings of the AAAI Conference on Artificial
  Intelligence}, volume~38, 12857--12865.

\bibitem[{Li et~al.(2024)Li, Song, Wu, Hu, Cheung, and Yao}]{10491302}
Li, S.; Song, L.; Wu, X.; Hu, Z.; Cheung, Y.-M.; and Yao, X. 2024.
\newblock Multi-Class Imbalance Classification Based on Data Distribution and
  Adaptive Weights.
\newblock \emph{IEEE Transactions on Knowledge and Data Engineering}, 36(10):
  5265--5279.

\bibitem[{Liu et~al.(2024)Liu, Jianye, Ma, and Xia}]{liuunlock}
Liu, J.; Jianye, H.; Ma, Y.; and Xia, S. 2024.
\newblock Unlock the Cognitive Generalization of Deep Reinforcement Learning
  via Granular Ball Representation.
\newblock In \emph{International Conference on Machine Learning}. PMLR.

\bibitem[{Liu and Dietterich(2012)}]{liu2012conditional}
Liu, L.; and Dietterich, T. 2012.
\newblock A conditional multinomial mixture model for superset label learning.
\newblock \emph{Advances in neural information processing systems}, 25.

\bibitem[{Liu et~al.(2021)Liu, Wang, Chen, Zhou, Zheng, and
  He}]{liu2021partial}
Liu, W.; Wang, L.; Chen, J.; Zhou, Y.; Zheng, R.; and He, J. 2021.
\newblock A partial label metric learning algorithm for class imbalanced data.
\newblock In \emph{Asian Conference on Machine Learning}, 1413--1428. PMLR.

\bibitem[{Lu et~al.(2023)Lu, Zhang, Han, Cheung, and Wang}]{Lu_2023_ICCV}
Lu, Y.; Zhang, Y.; Han, B.; Cheung, Y.-M.; and Wang, H. 2023.
\newblock Label-Noise Learning with Intrinsically Long-Tailed Data.
\newblock In \emph{Proceedings of the IEEE/CVF International Conference on
  Computer Vision (ICCV)}, 1369--1378.

\bibitem[{Lv et~al.(2020)Lv, Xu, Feng, Niu, Geng, and
  Sugiyama}]{lv2020progressive}
Lv, J.; Xu, M.; Feng, L.; Niu, G.; Geng, X.; and Sugiyama, M. 2020.
\newblock Progressive identification of true labels for partial-label learning.
\newblock In \emph{international conference on machine learning}, 6500--6510.
  PMLR.

\bibitem[{Lyu, Wu, and Feng(2022)}]{lyu2022deep}
Lyu, G.; Wu, Y.; and Feng, S. 2022.
\newblock Deep graph matching for partial label learning.
\newblock In \emph{Proceedings of the International Joint Conference on
  Artificial Intelligence}, 3306--3312.

\bibitem[{Nguyen and Caruana(2008)}]{nguyen2008classification}
Nguyen, N.; and Caruana, R. 2008.
\newblock Classification with partial labels.
\newblock In \emph{Proceedings of the 14th ACM SIGKDD international conference
  on Knowledge discovery and data mining}, 551--559.

\bibitem[{Qian et~al.(2024{\natexlab{a}})Qian, Li, Ye, Xia, Huang, and
  Ding}]{10539299}
Qian, W.; Li, Y.; Ye, Q.; Xia, S.; Huang, J.; and Ding, W. 2024{\natexlab{a}}.
\newblock Confidence-Induced Granular Partial Label Feature Selection via
  Dependency and Similarity.
\newblock \emph{IEEE Transactions on Knowledge and Data Engineering}, 36(11):
  5797--5810.

\bibitem[{Qian et~al.(2024{\natexlab{b}})Qian, Tu, Huang, Shu, and
  Cheung}]{10416802}
Qian, W.; Tu, Y.; Huang, J.; Shu, W.; and Cheung, Y.-M. 2024{\natexlab{b}}.
\newblock Partial Multilabel Learning Using Noise-Tolerant Broad Learning
  System With Label Enhancement and Dimensionality Reduction.
\newblock \emph{IEEE Transactions on Neural Networks and Learning Systems},
  1--15.

\bibitem[{Tang et~al.(2022)Tang, Tao, Qi, Liu, and Zhang}]{tang2022invariant}
Tang, K.; Tao, M.; Qi, J.; Liu, Z.; and Zhang, H. 2022.
\newblock Invariant feature learning for generalized long-tailed
  classification.
\newblock In \emph{European Conference on Computer Vision}, 709--726. Springer.

\bibitem[{Tian, Yu, and Fu(2023)}]{tian2023partial}
Tian, Y.; Yu, X.; and Fu, S. 2023.
\newblock Partial label learning: Taxonomy, analysis and outlook.
\newblock \emph{Neural Networks}.

\bibitem[{Wang, Li, and Zhang(2019)}]{wang2019adaptive}
Wang, D.-B.; Li, L.; and Zhang, M.-L. 2019.
\newblock Adaptive graph guided disambiguation for partial label learning.
\newblock In \emph{Proceedings of the 25th ACM SIGKDD International Conference
  on Knowledge Discovery \& Data Mining}, 83--91.

\bibitem[{Wang et~al.(2022{\natexlab{a}})Wang, Xia, Li, Mao, Feng, Chen, and
  Zhao}]{wang2022solar}
Wang, H.; Xia, M.; Li, Y.; Mao, Y.; Feng, L.; Chen, G.; and Zhao, J.
  2022{\natexlab{a}}.
\newblock Solar: Sinkhorn label refinery for imbalanced partial-label learning.
\newblock \emph{Advances in neural information processing systems}, 35:
  8104--8117.

\bibitem[{Wang et~al.(2022{\natexlab{b}})Wang, Xiao, Li, Feng, Niu, Chen, and
  Zhao}]{wang2022pico}
Wang, H.; Xiao, R.; Li, Y.; Feng, L.; Niu, G.; Chen, G.; and Zhao, J.
  2022{\natexlab{b}}.
\newblock Pico: Contrastive label disambiguation for partial label learning.
\newblock In \emph{International conference on learning representations}.

\bibitem[{Wang and Zhang(2018)}]{wang2018towards}
Wang, J.; and Zhang, M.-L. 2018.
\newblock Towards mitigating the class-imbalance problem for partial label
  learning.
\newblock In \emph{Proceedings of the 24th ACM SIGKDD International Conference
  on Knowledge Discovery \& Data Mining}, 2427--2436.

\bibitem[{Wen et~al.(2021)Wen, Cui, Hang, Liu, Wang, and
  Lin}]{wen2021leveraged}
Wen, H.; Cui, J.; Hang, H.; Liu, J.; Wang, Y.; and Lin, Z. 2021.
\newblock Leveraged weighted loss for partial label learning.
\newblock In \emph{International conference on machine learning}, 11091--11100.
  PMLR.

\bibitem[{Xia et~al.(2022)Xia, Dai, Wang, Gao, and Giem}]{xia2022efficient}
Xia, S.; Dai, X.; Wang, G.; Gao, X.; and Giem, E. 2022.
\newblock An efficient and adaptive granular-ball generation method in
  classification problem.
\newblock \emph{IEEE Transactions on Neural Networks and Learning Systems}.

\bibitem[{Xia et~al.(2024)Xia, Lian, Wang, Gao, Hu, and
  Shao}]{DBLP:journals/tkde/XiaLWGHS24}
Xia, S.; Lian, X.; Wang, G.; Gao, X.; Hu, Q.; and Shao, Y. 2024.
\newblock Granular-Ball Fuzzy Set and Its Implement in {SVM}.
\newblock \emph{{IEEE} Trans. Knowl. Data Eng.}, 36(11): 6293--6304.

\bibitem[{Xia et~al.(2019)Xia, Liu, Ding, Wang, Yu, and Luo}]{xia2019granular}
Xia, S.; Liu, Y.; Ding, X.; Wang, G.; Yu, H.; and Luo, Y. 2019.
\newblock Granular ball computing classifiers for efficient, scalable and
  robust learning.
\newblock \emph{Information Sciences}, 483: 136--152.

\bibitem[{Xia et~al.(2021)Xia, Zheng, Wang, Gao, and Wang}]{xia2021granular}
Xia, S.; Zheng, S.; Wang, G.; Gao, X.; and Wang, B. 2021.
\newblock Granular ball sampling for noisy label classification or imbalanced
  classification.
\newblock \emph{IEEE Transactions on Neural Networks and Learning Systems}.

\bibitem[{Xie et~al.(2024{\natexlab{a}})Xie, Hua, Xia, Cheng, Wang, and
  Gao}]{xie2024w}
Xie, J.; Hua, C.; Xia, S.; Cheng, Y.; Wang, G.; and Gao, X. 2024{\natexlab{a}}.
\newblock W-GBC: An Adaptive Weighted Clustering Method Based on Granular-Ball
  Structure.
\newblock In \emph{2024 IEEE 40th International Conference on Data Engineering
  (ICDE)}, 914--925. IEEE.

\bibitem[{Xie et~al.(2024{\natexlab{b}})Xie, Xiang, Xia, Jiang, Wang, and
  Gao}]{xie2024mgnr}
Xie, J.; Xiang, X.; Xia, S.; Jiang, L.; Wang, G.; and Gao, X.
  2024{\natexlab{b}}.
\newblock MGNR: A Multi-Granularity Neighbor Relationship and Its Application
  in KNN Classification and Clustering Methods.
\newblock \emph{IEEE Transactions on Pattern Analysis and Machine
  Intelligence}.

\bibitem[{Xie et~al.(2024{\natexlab{c}})Xie, Zhang, Luo, and
  Wang}]{xie2024three}
Xie, Q.; Zhang, Q.; Luo, N.; and Wang, G. 2024{\natexlab{c}}.
\newblock Three-Way Hybrid Sampling Using Granular Balls for Imbalanced
  Classification.
\newblock In \emph{International Joint Conference on Rough Sets}, 86--102.
  Springer.

\bibitem[{Xie et~al.(2024{\natexlab{d}})Xie, Zhang, Xia, Zhao, Wu, Wang, and
  Ding}]{DBLP:journals/tetci/XieZXZWWD24}
Xie, Q.; Zhang, Q.; Xia, S.; Zhao, F.; Wu, C.; Wang, G.; and Ding, W.
  2024{\natexlab{d}}.
\newblock {GBG++:} {A} Fast and Stable Granular Ball Generation Method for
  Classification.
\newblock \emph{{IEEE} Trans. Emerg. Top. Comput. Intell.}, 8(2): 2022--2036.

\bibitem[{Xu et~al.(2025)Xu, Qian, Cai, Shu, Huang, Cheung, and
  Ding}]{xu2025label}
Xu, F.; Qian, W.; Cai, X.; Shu, W.; Huang, J.; Cheung, Y.-M.; and Ding, W.
  2025.
\newblock Label Disambiguation-Based Feature Selection for Partial Multi-label
  Learning.
\newblock In \emph{International Conference on Pattern Recognition}, 265--279.
  Springer.

\bibitem[{Xu et~al.(2024)Xu, Lian, Liu, Chen, and Tao}]{xu2024pseudo}
Xu, M.; Lian, Z.; Liu, B.; Chen, Z.; and Tao, J. 2024.
\newblock Pseudo Labels Regularization for Imbalanced Partial-Label Learning.
\newblock In \emph{ICASSP 2024-2024 IEEE International Conference on Acoustics,
  Speech and Signal Processing (ICASSP)}, 6305--6309. IEEE.

\bibitem[{Xu et~al.(2021)Xu, Qiao, Geng, and Zhang}]{xu2021instance}
Xu, N.; Qiao, C.; Geng, X.; and Zhang, M.-L. 2021.
\newblock Instance-dependent partial label learning.
\newblock \emph{Advances in Neural Information Processing Systems}, 34:
  27119--27130.

\bibitem[{Zhang et~al.(2020)Zhang, Li, Jiang, and Tan}]{zhang2020learning}
Zhang, J.; Li, S.; Jiang, M.; and Tan, K.~C. 2020.
\newblock Learning from weakly labeled data based on manifold regularized
  sparse model.
\newblock \emph{IEEE Transactions on Cybernetics}, 52(5): 3841--3854.

\bibitem[{Zhang et~al.(2023)Zhang, Kang, Hooi, Yan, and Feng}]{zhang2023deep}
Zhang, Y.; Kang, B.; Hooi, B.; Yan, S.; and Feng, J. 2023.
\newblock Deep long-tailed learning: A survey.
\newblock \emph{IEEE Transactions on Pattern Analysis and Machine
  Intelligence}, 45(9): 10795--10816.

\end{thebibliography}

\end{document}